\documentclass[lettersize,journal]{IEEEtran}
\usepackage{amsmath,amsfonts}
\usepackage{algorithm, algorithmic}
\usepackage{array}
\usepackage[caption=false,font=normalsize,labelfont=sf,textfont=sf]{subfig}
\usepackage{textcomp}
\usepackage{stfloats}
\usepackage{url}
\usepackage{verbatim}
\usepackage{graphicx}
\usepackage{tabularx}
\usepackage{microtype}
\usepackage{multirow}
\usepackage[normalem]{ulem}
\usepackage[inkscapelatex=false]{svg}
\usepackage{hyperref}

\def\BibTeX{{\rm B\kern-.05em{\sc i\kern-.025em b}\kern-.08em
    T\kern-.1667em\lower.7ex\hbox{E}\kern-.125emX}}
\usepackage{balance}
\begin{document}
\title{Multi-Observer Vehicle Localization Case Study with Roadside Radar and Connected Vehicle Sensing}
\author{Aleksi Pippuri, Nilusha Jayawickrama, Risto Ojala
\thanks{Aleksi Pippuri (email: aleksi.pippuri@aalto.fi), (email: nilusha.jayawickrama@aalto.fi), and Risto Ojala (email: risto.j.ojala@aalto.fi) are with the Department of Mechanical Engineering, Aalto University, Espoo, Finland.  OpenAI ChatGPT 5.5 was used for editing and grammar enhancement. After using this tool/service, the author(s) reviewed and edited the content as needed and take(s) full responsibility for the content of the publication. This work was supported in part by the Finnish Centre for Economic Development, Transport and the Environment under the TECHBOOST project
(0124/05.02.09/2023A).}}

\markboth{}%
{}

\maketitle

\begin{abstract}
In modern intelligent transportation systems, it is essential to accurately estimate vehicle positions, especially in mixed traffic conditions where both connected and conventional vehicles coexist. Roadside infrastructure and connected vehicles can provide complementary observations of the same traffic scene, but real-world evidence on decision-level fusion between these sources remains limited. This paper proposes a multi-observer vehicle localization framework that fuses compact object-level detections from a static roadside radar and a dynamic LiDAR-equipped connected vehicle. We evaluate the framework with real-world data collected at an urban intersection in Helsinki, Finland, with a separately instrumented target vehicle used as the reference trajectory. Two extended Kalman filter based strategies for the localization task were benchmarked. The performance of the radar and LiDAR sensors were evaluated separately, and the two fusion strategies were explored under nominal sensing conditions, reduced LiDAR update rates, simulated LiDAR occlusions, and different target-vehicle motion states. The results show that, under full LiDAR availability, fusion performance is dominated by the LiDAR observations, while the less accurate and less consistent radar observations provide only limited additional improvement. Nevertheless, AEKF achieves small gains over the LiDAR-only baseline, and object-level connected vehicle observations remain useful when shared at reduced update rates. These findings indicate that decision-level fusion provides scenario-dependent benefits rather than automatic improvement over a strong single-sensor baseline. We release the dataset and implementation on Github to support further research: https://github.com/AppuriAalto/multi-observer-vehicle-tracking
\end{abstract}


\section{Introduction}

\IEEEPARstart{C}{onnected} vehicles (CVs), together with intelligent transportation systems (ITS), aim to revolutionize the near future of mobility by increasing safety, reducing travel times, and lowering emissions \cite{Olia2016}. These ambitions rely on the coordinated operation of multiple technologies, applications, and participants, including both individual drivers and infrastructure administrators.

An integral technological enabler of ITS is vehicle-to-everything (V2X) communication, which allows CVs to exchange information directly with each other (V2V) and with roadside infrastructure (V2I). V2I communications can enhance existing systems such as adaptive traffic signal control \cite{Ghoul2021}, while V2V enables new applications such as cooperative collision avoidance \cite{ETSITR102638}. Together, these capabilities are realized in vehicular ad-hoc networks (VANETs), where on-board units in vehicles communicate with roadside units connected to edge computing units. Such networks allow local traffic information, road user detections, and other cooperative awareness messages to be shared dynamically between participants, either directly or via multi-hop communication.

Reliable ITS operation requires accurate real-time vehicle position information. However, widespread adoption of V2X remains years away, and traffic will continue to include both connected and conventional vehicles. This is particularly true in Europe, where the average passenger car age was 12.3 years in 2024 according to the European Automobile Manufacturers' Association \cite{ACEA2024}. In mixed-traffic,  ITS must infer information about conventional vehicles based on perception and communication of participating CVs and infrastructure.

\begin{figure}[t]
    \centering
    \includegraphics[width=\columnwidth]{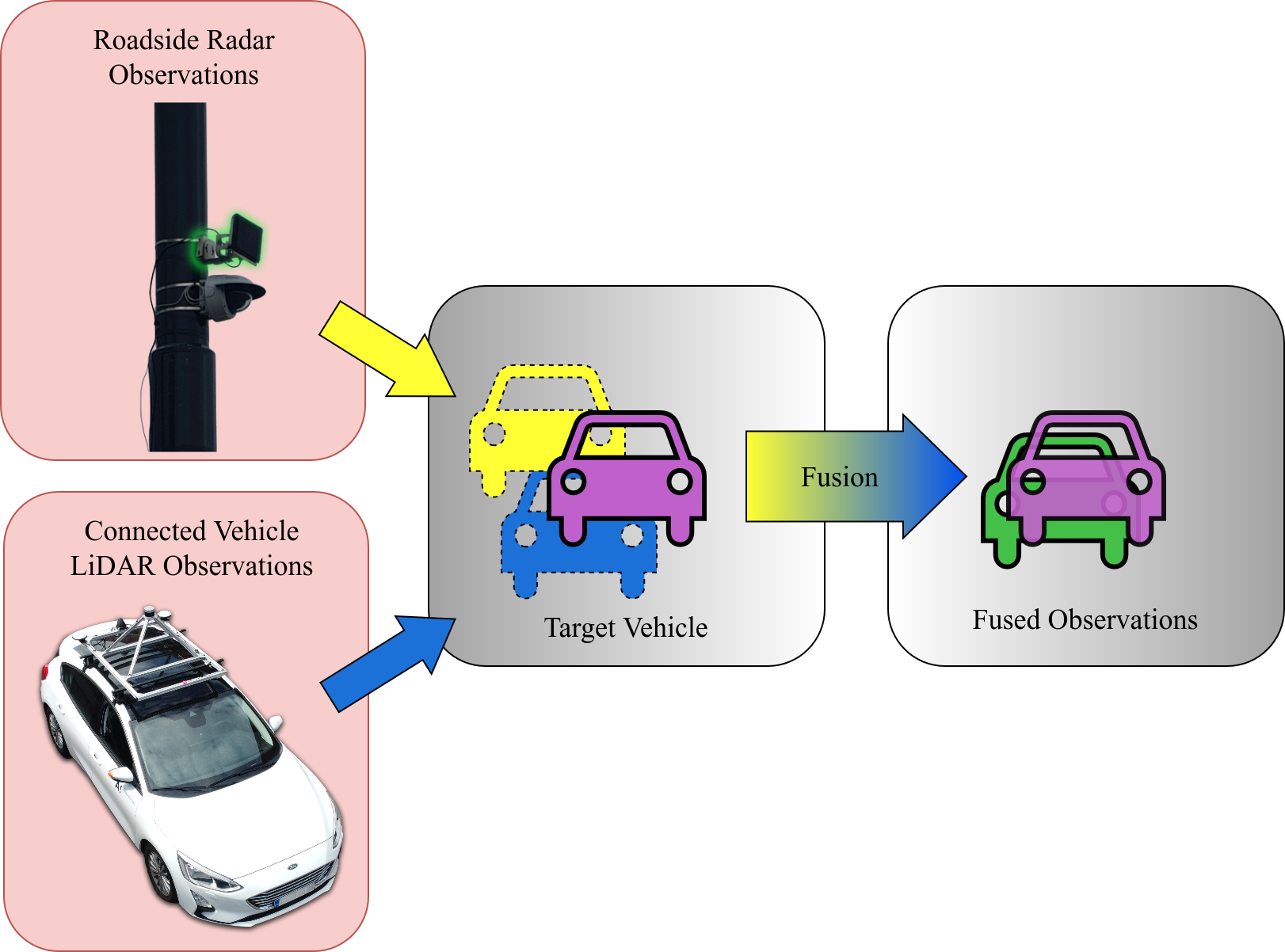}
    \caption{The proposed framework takes as input vehicle detections from a roadside radar and a CV-equipped LiDAR. The detections are then fused and evaluated against a known reference track from a target vehicle.}
    \label{fig:intro_flowchart}
\end{figure}

Beyond penetration rates, ITS performance is also constrained by environmental and technological factors. These include environmental conditions such as weather, occlusions, and obstructions, as well as technological limitations including heterogeneous sensor modalities, incompatible data formats, and channel congestion \cite{Arnold2019}. 

To address these challenges, data fusion (DF) techniques are increasingly applied. DF combines information from multiple sources to enhance accuracy, precision, and reliability compared to single-sensor observations. Fusion methods can be categorized as data-level (combining raw data), feature-level (combining extracted features), or decision-level (fusing processed attributes such as object locations or tracks) \cite{Kashinath2021,Schmitt2016}. In ITS, data-level fusion is often infeasible due to bandwidth demands and sensor heterogeneity, while feature-level fusion is mainly suited to homogeneous sensors. Decision-level fusion is therefore the most practical option for heterogeneous and distributed detection sources.

We implement and evaluate a decision-level data fusion framework for vehicle localization with multiple observers in a real-world mixed-traffic urban scenario as described in Figure~\ref{fig:intro_flowchart}. Specifically, the proposed pipeline fuses vehicle detections from heterogeneous observers, and sequential extended Kalman filter (SEKF)- and averaged EKF (AEKF)-based fusion variants are benchmarked for state estimation. The architecture allows observations from both static infrastructure sensor and dynamic CV platforms to be combined depending on availability. The framework is validated using real-world measurements from a roadside radar and a CV equipped with LiDAR and GNSS/INS. Moreover, we analyze the effects of reduced CV observations to identify when fusion participants are able to contribute to the performance of the filter.

\noindent\textbf{The main contributions of this paper are as follows:}
\begin{enumerate}
    \item We present a decision-level multi-observer vehicle tracking framework that benchmarks SEKF- and AEKF-based fusion variants using heterogeneous observations from roadside radar and a LiDAR/GNSS-equipped CV.

    \item We evaluate the framework with real-world data under varying observation sources, target motion states, and CV observation rates, showing that CV observations can benefit ITS even at modest update rates.

    \item We release the software implementation of the proposed fusion and tracking pipeline and the corresponding real-world validation data as open source on GitHub to support reproducibility and further research.

\end{enumerate}

\section{Related Work}

\subsection{Cooperative Perception and Decision-level Fusion in ITS}

Decision-level fusion, for perception tasks in ITS, focuses on communicating compact object-level detections or tracks instead of raw sensor streams or dense feature representations. As a result, it is more appealing for heterogeneous roadside and vehicle-based observers than data-level or feature-level fusion, which are more constrained by bandwidth, modality compatibility, and shared feature representations \cite{Kashinath2021,Schmitt2016}. However, real-world studies of decision-level fusion across heterogeneous infrastructure and connected-vehicle observers are limited.

While decision-level fusion in collaborative perception context has been extensively studied in simulations, real-world experimentation remains limited. Chen \cite{Chen2024} proposed a method for sharing LiDAR-based static sensor information with vehicles to improve situational awareness within the Carla simulator. Malinvero et al. \cite{Malinvero2020} developed a cooperative collision avoidance system that prevented most potential collisions in simulation. However, their approach required high penetration rates of CVs, limiting near-future applicability. Similarly, Dong et al. \cite{Dong2020} investigated real-time tracking but restricted their work to camera-based solutions, while Feng et al. \cite{Feng2015} explored estimation of non-connected vehicles using CV data. Chen \cite{Chen2023} further researched data-level fusion between camera and radar, but scalability across multiple modalities was not addressed. More recently, Liu et al. \cite{Liu2023} introduced adaptive factors for covariance and measurement matrices based on detection confidences, along with a confidence decay mechanism when tracks were unmatched. Their approach showed improvements over state-of-the-art methods, but was designed for onboard fusion between LiDAR and camera sensors in a single vehicle. A follow-up by Liu et al. in 2025 \cite{Liu2025} extended this line of work but remained vehicle-centric. By contrast, the method proposed in this paper focuses on scalable, vehicle-assisted decision-level fusion across heterogeneous observers. 

Publicly available datasets enabling collaborative perception in traffic environments are either in simulated environments or, when in the real world, commonly use camera- and LiDAR-based sensing setups. For instance, V2XSim \cite{Li2022} and DOLPHINS \cite{Mao2023} are simulated in CARLA, where they provide collaborative perception scenarios with visual and LiDAR-based sensing. Real-world datasets such as V2V4Real \cite{Xu2023}, TUMTraf-V2X \cite{Zimmer2024}, and V2X-Seq \cite{Yu2023} provide valuable V2V or V2I perception data, typically using combinations of cameras, LiDARs, and localization sensors such as GNSS/IMU.

Some hybrid cooperative fusion approaches that combine onboard sensing with external infrastructure data have also emerged. Gabb et al. \cite{Gabb2019} proposed a heterogeneous fusion architecture incorporating track-level information from a multi-access edge computing server alongside onboard measurements. Related V2I studies addressing vulnerable road users and mixed traffic environments \cite{George2025}, as well as V2V-enabled Kalman-filter-based trajectory estimation for cooperative collision avoidance \cite{Zhang2017}. These works further demonstrate the growing interest in cooperative perception, although large-scale real-world validation with heterogeneous infrastructure and CV observations remains limited.

\subsection{Kalman filter-based Vehicle Tracking and Sensor Fusion}

Kalman filter-based modules are popular for vehicle tracking because they can be tuned to output recursive state estimation with inexpensive computational cost and also to integrate measurement data to the module as and when they become available \cite{khodarahmi2023review}. When a tracking problem becomes nonlinear, prior works have highlighted the potential of leveraging an EKF in order to estimate the state of vehicles by using position and velocity-related measurements as input \cite{seo2025dynamic}. Other approaches for improving the robustness under temporal challenges, such as varying measurement noise, sensor degradation, and complex target motion, include adaptive KF \cite{Shafik2025}, covariance matching strategies \cite{Brown1985}, and interacting multiple model (IMM) filters \cite{Sumithra2021}. Yet, despite high tracking flexibility, evidence from their previous usages indicate that they are typically suitable for single-platform scenarios. 

Several onboard fusion strategies have been investigated in previous research. Meng et al. \cite{Meng2025} introduced a Sage-Husa adaptive fading EKF that emphasizes recent measurements through a fading factor. Yoon et al. \cite{Yoon2023} applied an IMM framework in which each onboard sensor independently tracks surrounding vehicles while dynamically switching motion models and estimating sensor reliability. Multi-camera fusion with adaptive covariance inflation has also been explored to prevent filter divergence and improve sensor weighting \cite{Li2017}. Although effective, these approaches also remain primarily vehicle-centric because they focus on fusing sensors mounted on a single ego vehicle.

Another form of KF-based estimation is SEKF, which incorporates measurements from multiple sensors one at a time as they arrive, without needing any synchronized measurements from all the sensors. Hence, SEKF is widely adopted in scenarios comprising distributed and multi-rate sensing scenarios \cite{li2019asynchronous}\cite{li2025sequential}. Thus, prior works show that this is especially beneficial for V2X since roadside infrastructure sensors and CV sensors tend to \textbf{1)} operate with different update rates, \textbf{2)} have intermittent detections, and \textbf{3)} provide observations of varying uncertainties. 

Further, averaging-based Kalman filter (AEKF) modules for fusion have also been designed for vehicle tracking in scenarios where multiple sensor observations for the same target are available  \cite{li2025construction}. Specifically, such approaches typically combine the available measurements prior to the filter update step by getting the average of position estimates from multiple observers. AEKF is comparatively one of the most computationally low-cost KF-based techniques, but previous work suggests that it could be sensitive to differences in factors such as sensor accuracy, update rate, and detection confidence \cite{mohammadisarab2025fusion}. Yet, it stands as one of the most useful baselines to assess if sequential fusion practically offers more advantages over simpler decision-level aggregation. 

In addition, two-stage adaptive Kalman filters have also been used for traffic signal control applications to estimate turning movements and queue lengths by fusing probe vehicle and fixed-sensor data \cite{Wang2025}.

\subsection{Research gap}

In summary, prior research has shown the potential of collaborative perception and filtering-based vehicle tracking in ITS, yet notable gaps remain. Particularly, most works are based on simulated environments and emphasize camera- and LiDAR-based sensing. In contrast, roadside radar is widely used in traffic monitoring and adaptive signal control, is comparatively affordable, and can operate under varying lighting and weather conditions. Despite this practical relevance, limited real-world evidence is available of \textbf{1)} decision-level fusion between static roadside radar and dynamic CV observations, and \textbf{2)} evidence as to when roadside radar fusion improves vehicle tracking via a CV perception architecture. This work contributes to filling these gaps by implementing and experimentally evaluating EKF-based decision-level fusion using heterogeneous observations from infrastructure-based radar and CV-based LiDAR in a real-world urban intersection.

\section{Methods}

\subsection{Problem Statement}

This work addresses the problem of vehicle localization using detections from a roadside radar and a CV-mounted LiDAR. Both observers provide two-dimensional vehicle detections in a common radar-centric frame when observations are available. The task is to associate asynchronous radar and LiDAR detections with vehicle identities and to estimate the vehicle trajectories over time. The objective is to track each vehicle with high localization accuracy while maintaining correct target identity through missed detections, asynchronous observations, and varying sensor availability. The target vehicle GNSS/INS trajectory is used only as a reference for calibration and evaluation.

\subsection{Tracking and Fusion Framework}

\begin{figure*}[t]
    \centering
    \includegraphics[width=0.95\textwidth]{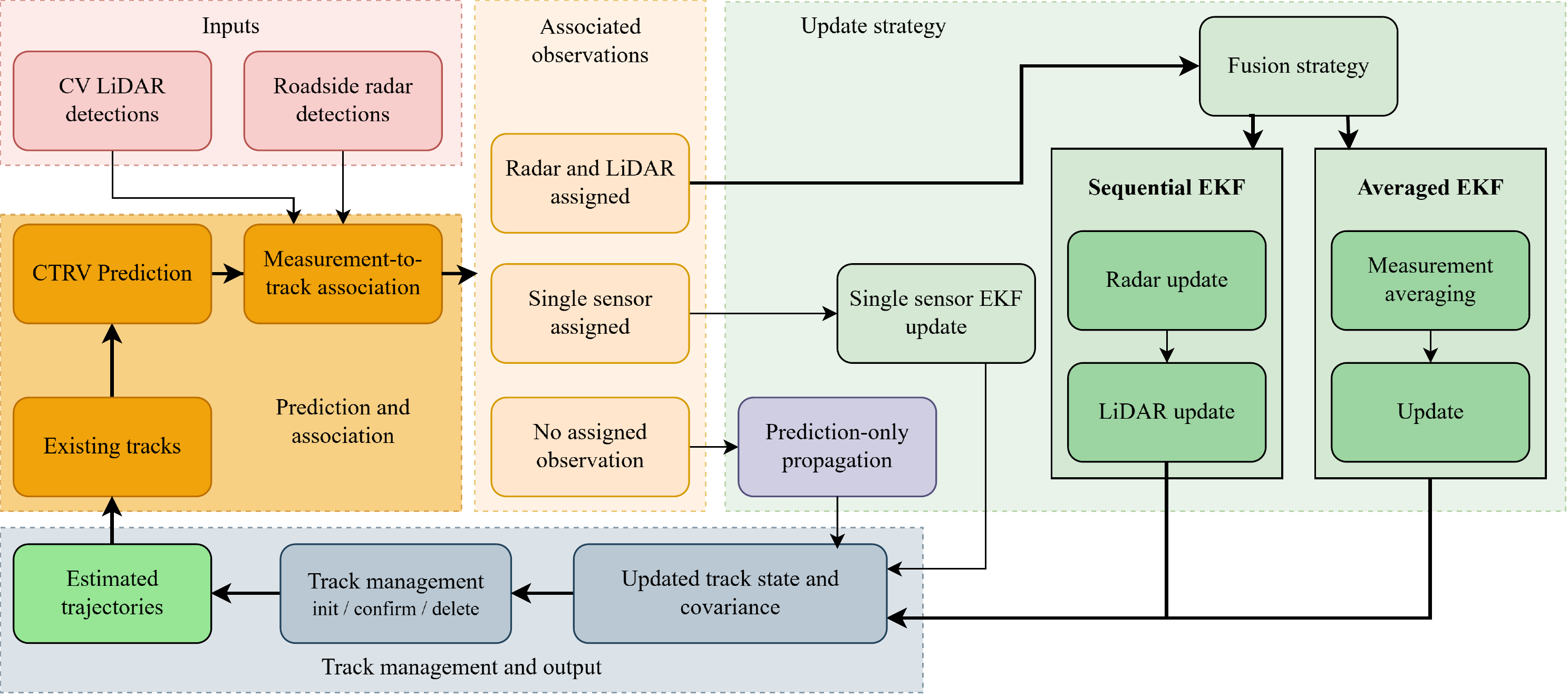}
    \caption{Proposed tracking and fusion framework. Existing tracks are propagated using the CTRV model and associated with radar and LiDAR detections. Depending on observation availability, the framework applies sequential updating, averaged updating, single-sensor updating, or prediction-only propagation. Track management handles initialization, confirmation, deletion, and trajectory output.}
    \label{fig:main_method_flowchart}
\end{figure*}

Figure~\ref{fig:main_method_flowchart} summarizes the full tracking and fusion pipeline. The CTRV prediction and measurement-to-track association stages are described in Sections~\ref{measurement model} and~\ref{measurement-to-track association} respectively. The observation availability logic determines whether the track receives radar and LiDAR observations, a single observation, or no observation. The corresponding update strategies are described in Sections~\ref{sequential ekf update} and~\ref{averaged ekf update}, while the single-sensor and prediction-only cases follow directly from the same EKF prediction and update formulation. Finally, track initialization, confirmation, deletion, and trajectory output are described in Section~\ref{track management}.

\subsubsection{State Estimation and Motion Model} \label{state estimation and motion model}

The states of the detected vehicles are estimated using an EKF based on the constant turn rate and velocity (CTRV) motion model. The system operates in the radar-centric 2D coordinate frame. The state vector is defined as

\begin{equation}
\mathbf{x} =
\begin{bmatrix}
p_x & p_y & v & \psi & \dot{\psi}
\end{bmatrix}^T
\end{equation}
where $p_x$ and $p_y$ denote position, $v$ is speed, $\psi$ is heading, and $\dot{\psi}$ is yaw rate.

The CTRV model is used for track propagation, assuming constant velocity and constant turn rate between consecutive updates. The EKF prediction step propagates the state with the CTRV motion model and updates the covariance using the corresponding Jacobian $\mathbf{F}_k$ and process noise covariance $\mathbf{Q}_k$. A straight-motion approximation is used when the yaw rate approaches zero, i.e., $|\dot{\psi}| \approx 0$.

\subsubsection{Measurement Model} \label{measurement model}

Both radar and LiDAR are used as position-only measurements in the
radar-centric coordinate frame. Radar speed and bearing are used only when initializing new radar-born tracks, where available, to provide an initial speed and heading estimate for the CTRV state. The regular measurement update therefore remains consistent across sensors, with both radar and LiDAR observations modeled as two-dimensional localizations.

For sensor $s \in \{r,l\}$, the measurement model is

\begin{subequations}
\label{eq:measurement_model}
\noindent
\begin{minipage}[t]{0.48\columnwidth}
\begin{equation}
\mathbf{z}_{s,k}
=
\mathbf{H}\mathbf{x}_{k}
+
\mathbf{v}_{s,k}
\label{eq:measurement_model_observation}
\end{equation}
\end{minipage}
\hfill
\begin{minipage}[t]{0.46\columnwidth}
\begin{equation}
\mathbf{v}_{s,k}
\sim
\mathcal{N}(0,\mathbf{R}_s)
\label{eq:measurement_model_noise}
\end{equation}
\end{minipage}
\end{subequations}
\vspace{0.8\baselineskip}

where

\begin{subequations}
\label{eq:measurement_matrices}
\noindent
\begin{minipage}[t]{0.48\columnwidth}
\begin{equation}
\setlength{\arraycolsep}{3pt}
\mathbf{H}
=
\begin{bmatrix}
1 & 0 & 0 & 0 & 0 \\
0 & 1 & 0 & 0 & 0
\end{bmatrix}
\label{eq:measurement_matrix_H}
\end{equation}
\end{minipage}
\hfill
\begin{minipage}[t]{0.46\columnwidth}
\begin{equation}
\mathbf{R}_s
=
\mathrm{diag}(\sigma_s^2,\sigma_s^2)
\label{eq:measurement_covariance_Rs}
\end{equation}
\end{minipage}
\end{subequations}

\vspace{0.8\baselineskip}

Here, $\sigma_s$ denotes the position measurement standard deviation of sensor $s$,
with $s=r$ for radar and $s=l$ for LiDAR.

\subsubsection{Sequential EKF Update} \label{sequential ekf update}

For each associated measurement, the state and covariance are updated using the standard EKF correction step with sensor-specific measurement covariance $\mathbf{R}_s$.

In the SEKF method, when both radar and LiDAR measurements are available at a given time step, updates are applied sequentially rather than jointly. Let $\mathbf{z}_r$ and $\mathbf{z}_l$ denote radar and LiDAR measurements, respectively. The update proceeds as:

\begin{subequations}
\label{eq:sekf_update}
\begin{align}
(\mathbf{x}, \mathbf{P}) &\leftarrow 
\operatorname{EKFUpdate}(\mathbf{x}, \mathbf{P}, \mathbf{z}_r, \mathbf{R}_r),
\label{eq:sekf_update_radar}
\\
(\mathbf{x}, \mathbf{P}) &\leftarrow 
\operatorname{EKFUpdate}(\mathbf{x}, \mathbf{P}, \mathbf{z}_l, \mathbf{R}_l).
\label{eq:sekf_update_lidar}
\end{align}
\end{subequations}

The update order is configurable but was kept fixed during execution as radar first followed by LiDAR. Sequential updating avoids the need for explicit joint measurement modeling and allows sensor-specific uncertainty handling, while preserving modularity in regards to observation sources.

\subsubsection{Averaged EKF Update} \label{averaged ekf update}

As an alternative to sequential updating, an AEKF variant is implemented in which radar and LiDAR measurements assigned to the same track are first combined into a single equivalent position measurement. Let $\mathbf{z}_r$ and $\mathbf{z}_l$ denote the associated radar and LiDAR position measurements, with corresponding measurement covariance matrices $\mathbf{R}_r$ and $\mathbf{R}_l$. Assuming independent measurement errors, the fused measurement covariance and fused position measurement are computed as:

\begin{subequations}
\label{eq:aekf_measurement_average}
\begin{align}
\bar{\mathbf{R}} &=
\left(\mathbf{R}_r^{-1} + \mathbf{R}_l^{-1}\right)^{-1},
\label{eq:aekf_average_covariance}
\\
\bar{\mathbf{z}} &=
\bar{\mathbf{R}}
\left(\mathbf{R}_r^{-1}\mathbf{z}_r + \mathbf{R}_l^{-1}\mathbf{z}_l\right).
\label{eq:aekf_average_measurement}
\end{align}
\end{subequations}

The EKF correction is then applied once using the averaged measurement:

\begin{equation}
(\mathbf{x}, \mathbf{P}) \leftarrow 
\operatorname{EKFUpdate}(\mathbf{x}, \mathbf{P}, \bar{\mathbf{z}}, \bar{\mathbf{R}}).
\label{eq:aekf_update}
\end{equation}

Compared with the sequential EKF, the AEKF avoids dependence on update order and represents simultaneous radar-LiDAR observations as a single covariance-weighted position estimate.

\subsubsection{Handling of Partial and Missing Observations}

The above methods are designed to produce predictions for existing tracks even if no measurements can be associated with them. This was done to enable the system to handle asynchronous and partially available observations. 

If either of the observers does not to produce a measurement, the EKF produces a single-sensor update using the corresponding sensor covariance. In the SEKF case, missing sensor measurements are skipped in the configured update sequence. In the AEKF case, measurement averaging is performed only when both radar and LiDAR measurements are assigned to the same track; otherwise the update reduces to a radar-only or LiDAR-only correction.

\subsection{Data Association} \label{data association}

\subsubsection{Measurement-to-Track Association} \label{measurement-to-track association}

Radar and LiDAR measurements are matched with existing tracks using gated nearest-neighbour association. Each active track is first predicted to the current radar time step. For a measurement $\mathbf{z}_{i,s}$ from sensor $s \in \{r,l\}$ and track $j$, the innovation is computed as

\begin{equation}
\boldsymbol{\nu}_{ij,s}
=
\mathbf{z}_{i,s}
-
\mathbf{H}\hat{\mathbf{x}}^{-}_{j}.
\end{equation}

where $\hat{\mathbf{x}}^{-}_{j}$ is the predicted state of track $j$, and
$\mathbf{H}$ maps the CTRV state to the measured position components. The corresponding innovation covariance is

\begin{equation}
\mathbf{S}_{j,s}
=
\mathbf{H}\mathbf{P}^{-}_{j}\mathbf{H}^{T}
+
\mathbf{R}_{s}.
\end{equation}

where $\mathbf{P}^{-}_{j}$ is the predicted covariance of track $j$, and
$\mathbf{R}_{s}$ is the sensor-specific position measurement covariance. The association distance is then given by the squared Mahalanobis distance

\begin{equation}
d^{2}_{ij,s}
=
\boldsymbol{\nu}_{ij,s}^{T}
\mathbf{S}_{j,s}^{-1}
\boldsymbol{\nu}_{ij,s}.
\end{equation}

A measurement--track pair is accepted only if

\begin{equation}
d^{2}_{ij,s} \leq \gamma_{\mathrm{assoc}}, 
\qquad \gamma_{\mathrm{assoc}} = 5.
\end{equation}

Candidate pairs passing the gate are sorted by Mahalanobis distance and assigned under a one-to-one constraint, so that each measurement and each track can be used at most once per association pass.

\subsubsection{Track Management} \label{track management}

At each radar time step, all existing tracks are first propagated with the CTRV prediction model and their missed-observation counters are incremented. Measurements are associated with existing tracks using the gated nearest-neighbor procedure described above. For associated tracks, the EKF update is applied, the hit counter is incremented, and the missed counter is reset to zero.

Measurements that remain unassigned after association are used to initialize new tentative tracks. The initial state is formed from the measured two-dimensional position. For radar-initialized tracks, the reported radar speed and bearing are additionally used to initialize the CTRV speed and heading when the speed lies within the configured valid range. The initial covariance is defined by the configured prior variances for position, speed, yaw, and yaw rate. 

A tentative track is promoted to a confirmed track once its hit counter reaches the confirmation threshold. In the final configuration, this threshold is set to one hit, so tracks are confirmed immediately after a valid measurement. This setting was chosen to avoid introducing a confirmation-delay bias in the evaluation, since stricter confirmation would disproportionately penalize methods with lower observation availability.

Confirmed tracks are retained through temporary observation gaps and are deleted only when the missed-observation counter exceeds the configured maximum. With the final configuration, tracks are removed after more than 50 consecutive missed radar-frame updates, corresponding to approximately $5~\mathrm{s}$ for the nominal radar update rate.

When no measurement is associated with an existing track at a given update step,
no correction is applied and the track remains prediction-only for that frame.
This allows the tracker to tolerate temporary occlusions, intermittent LiDAR
availability, and radar detections that fail the association gate. Only confirmed tracks are included in the output trajectories used for evaluation.

\subsection{Experimental Setup}

\begin{figure}[ht]
    \centering
    \includegraphics[width=\columnwidth]{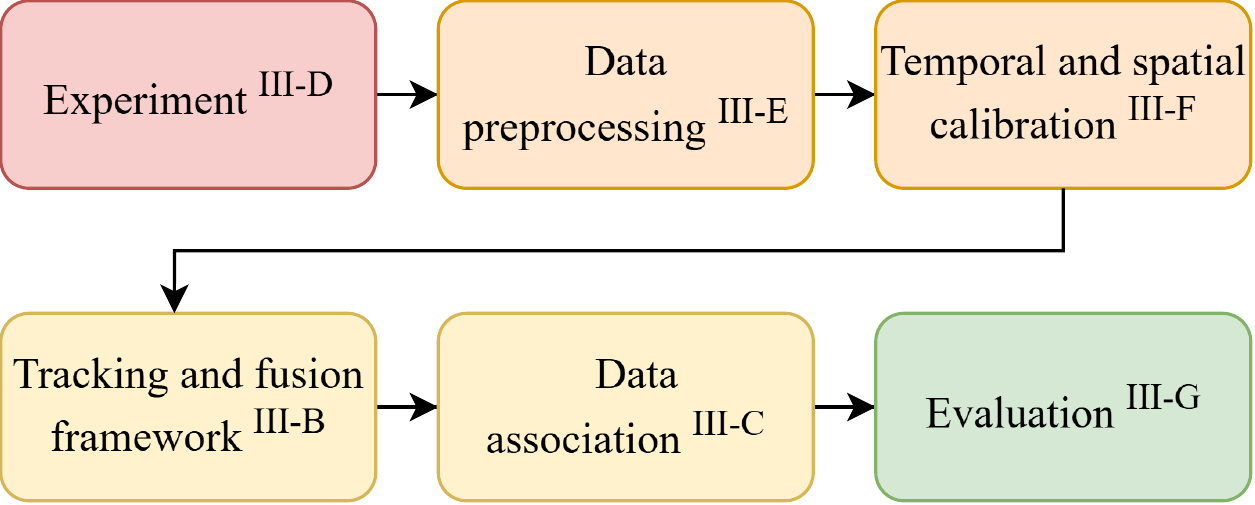}
    \caption{Overview of the experimental workflow. Data collected in the experiment are processed, calibrated, and then used in the proposed fusion framework. Finally the resulting tracks are evaluated against the target vehicle's reference trajectory.}
    \label{fig:Methods process}
\end{figure}

The experimental workflow, shown in Figure~\ref{fig:Methods process}, consists of data collection, pre-processing, temporal and spatial calibration, fusion, and evaluation.

\subsubsection{Experiment Design} \label{Experiment design}

The experiment was conducted in live urban traffic at an intersection in Helsinki, Finland. The sensing setup consisted of a permanently installed roadside radar mounted on a lamp post, a CV equipped with LiDAR and GNSS/INS unit, and an independent target vehicle equipped with a GNSS/INS unit. The radar overlooked a divided road section with three lanes in the approach direction and four lanes in the departure direction. The roadside radar and the CV observed surrounding traffic, while the target vehicle trajectory was used only as the independent reference for evaluation.

A total of 15 runs were collected during the same measurement session. Each run was divided into a departure and an approach segment, separated by a section outside the radar field of view. The segmentation enabled both vehicle arrangements to be evaluated: During the departure segment the target vehicle was driven in front of the CV and vice-versa during the approach segment.

\begin{figure}[h]
    \centering
    \includegraphics[width=\columnwidth]{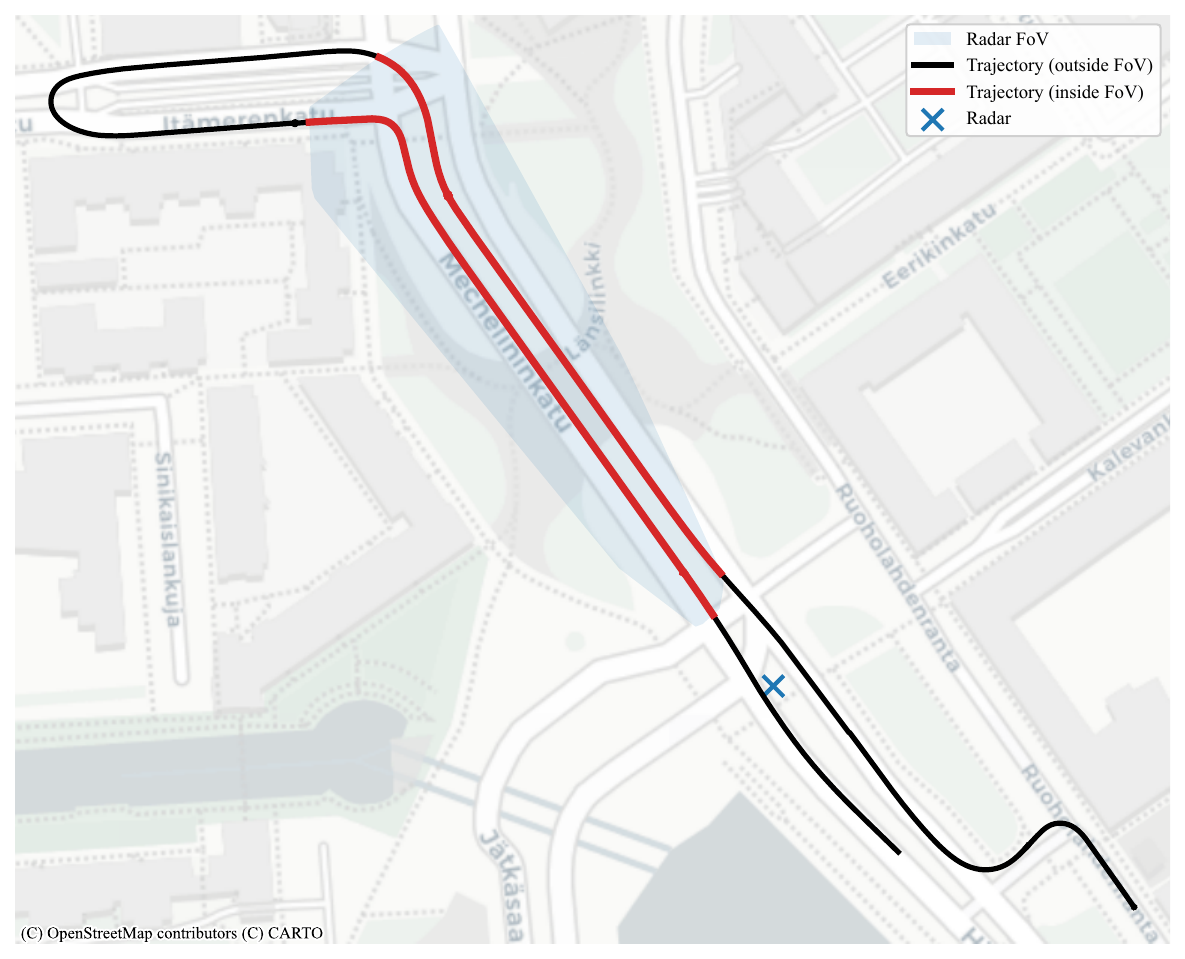}
    \caption{Target vehicle trajectory as well as the radar location shown. The segments of the route inside the radar FoV is highlighted, the right-side track being the depart segment and the left one approach. Map provided by OpenStreetMap \cite{OpenStreetMap2026} and CARTO \cite{CARTO2026}}.
    \label{fig:route_fov_map}
\end{figure}

The route shown in Figure~\ref{fig:route_fov_map} was repeated as consistently as possible within normal traffic conditions. The drivers were instructed to follow normal traffic behavior, use the same lanes across runs, maintain a typical following distance, accelerate moderately, and remain within the local speed limit of $40~\mathrm{km/h}$.

The 15 runs were split into five calibration and ten evaluation runs. However, due to file corruption one radar recording from the depart segment of the evaluation set is missing so the total becomes 10 calibration segments and 19 evaluation segments.

\subsubsection{System Configuration}

The experimental setup consisted of a) the CV equipped with a Velodyne VLP-32C LiDAR producing point clouds at 10 Hz and a Novatel PwrPak7D-E2 GNSS/INS unit with dual antennas producing accurate localization data at 50 Hz. These were connected with pulse per second signal to enable intra-vehicle time synchronization. b) the roadside radar which is a SmartMicro UMRR-11 type 44 producing vehicle locations and velocities at 10 Hz. And c) the target vehicle equipped with a Novatel PwrPak7D-E2 GNSS unit with dual antennas also producing localization data at 50 Hz.

\subsection{Data Preprocessing}

\begin{figure}[ht]
    \centering
    \includegraphics[width=\columnwidth]{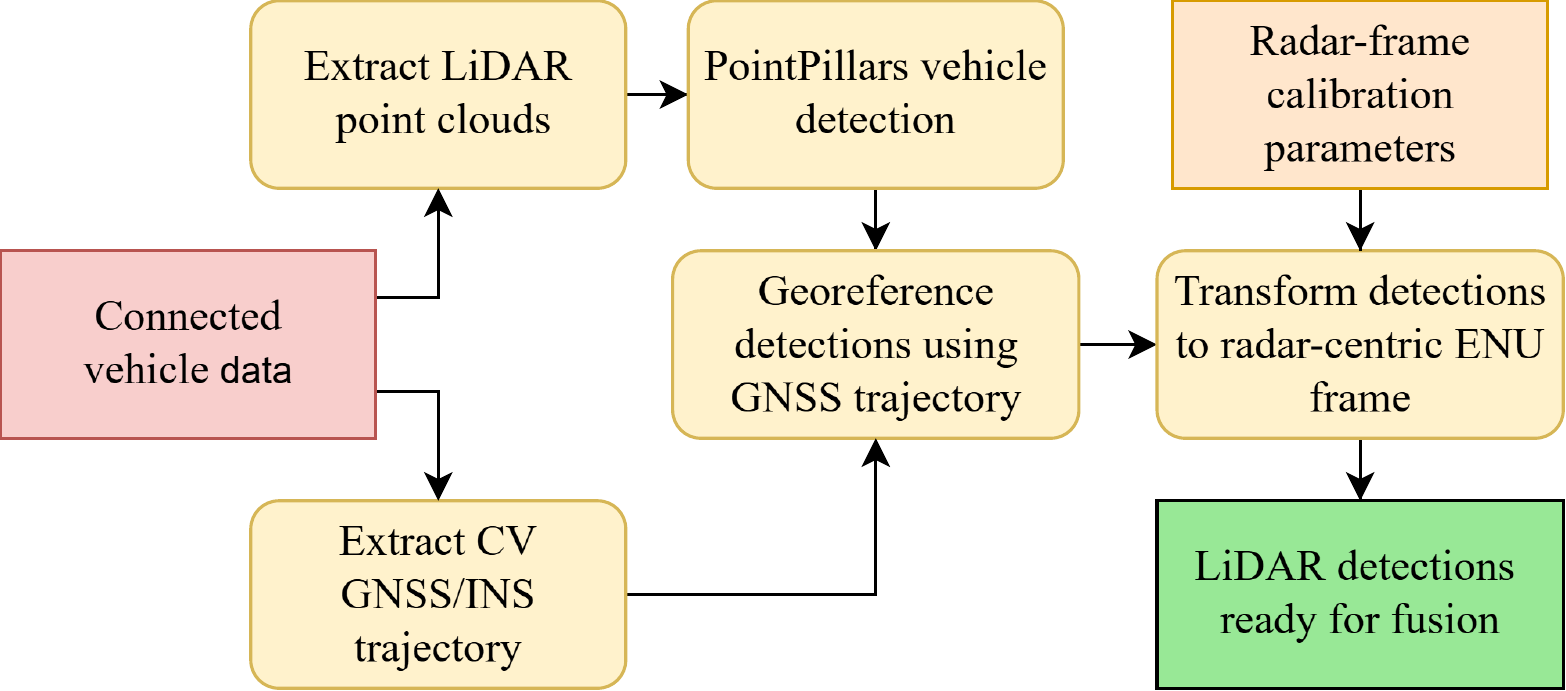}
    \caption{Connected vehicle preprocessing pipeline. LiDAR-based vehicle detections are georeferenced using the CV GNSS trajectory and transformed into the common radar-centric ENU frame for fusion.}
    \label{fig:CV preprocessing}
\end{figure}

Each sensing node recorded data locally, after which the datasets were aggregated for offline pre-processing and evaluation. In the CV preprocessing step, as shown in Figure~\ref{fig:CV preprocessing}, the point clouds were fed to a pretrained PointPillars \cite{lang2019pointpillars} object detection model running via OpenPCdet \cite{openpcdet2020} library.

The vehicle detections were then filtered with a confidence threshold of $0.2$ to remove noise. These were then transformed to geodetic WGS84 coordinates using the GNSS localization. Finally, the detections were transformed into radar-centric Cartesian ENU coordinates for the purpose of homogenizing the detections with the radar detections. This enabled the use of them in the DF methods and as a separate reference comparison.

\begin{figure}[ht]
    \centering
    \includegraphics[width=\columnwidth]{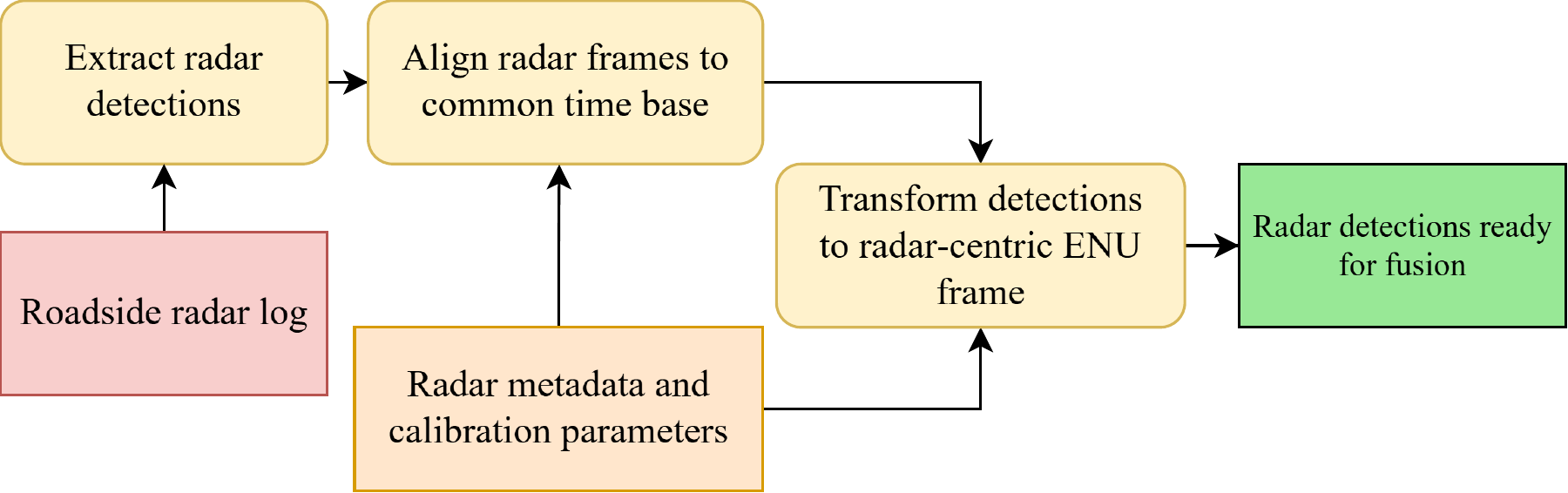}
    \caption{Roadside radar preprocessing pipeline. Radar-based vehicle detections are extracted, aligned to the CV time base, and transformed into the radar-centric ENU frame for fusion.}
    \label{fig:Radar preprocessing pipeline}
\end{figure}

The preprocessing procedure for the roadside radar is described in Figure~\ref{fig:Radar preprocessing pipeline}. The output detections obtained from the radar were in geodesic WGS84 coordinates, and so was the GNSS data coming from the target vehicle. Both were transformed into radar-centric ENU (east, north, up) coordinates.  The synchronization and geometric calibration between the sensing nodes are explained in section \ref{calibration}. 

\subsection{Temporal and Spatial Calibration} \label{calibration}

Since the experiment combines independent measurements from a CV, a roadside radar, and a target vehicle, all data streams must be expressed in a common temporal and spatial reference before fusion and later evaluation. The common time base used throughout the processing pipeline is denoted by $t_\mathrm{cv}$ and is tied to the CV GNSS time.

The radar spatial transform and clock offset were estimated by comparing radar detections with CV LiDAR detections and the GNSS trajectory in a radar-centric local ENU frame. The calibration method estimated a shared spatial and temporal correction over the calibration dataset with a parameter vector:

\begin{equation}
\boldsymbol{\theta}
=
\begin{bmatrix}
\Delta x_{\mathrm{cal}} &
\Delta y_{\mathrm{cal}} &
\psi_{\mathrm{cal}} &
\Delta t_{\mathrm{radar}}
\end{bmatrix}^{\mathsf{T}}.
\label{eq:calibration_parameter_vector}
\end{equation}

For each candidate parameter vector, radar and LiDAR detections were temporally matched within a fixed window and spatially associated using Hungarian assignment with a $d_{\mathrm{gate}}=4.0$~m gate. This value was selected to tolerate the initial radar spatial and temporal misalignment. The calibration residual was:

\begin{equation}
e_k(\boldsymbol{\theta})
=
\left\|
\mathbf{p}^{\mathrm{cal}}_{r,k}(\boldsymbol{\theta})
-
\mathbf{p}^{\mathrm{ref}}_{\ell,k}
\right\|_2 .
\label{eq:calibration_residual}
\end{equation}

where $\mathbf{p}^{\mathrm{cal}}_{r,k}$ is the calibrated radar detection position and $\mathbf{p}^{\mathrm{ref}}_{\ell,k}$ is the LiDAR/CV reference position obtained from its global WGS84 coordinate in the same local ENU frame.

As the objective, the residuals were summarized for each segment using a scalar objective combining trimmed RMSE (root mean square error), median and 90th percentile error, a match-shortage penalty and a weak prior:

\begin{equation}
\begin{aligned}
J_s(\boldsymbol{\theta})
=&\ \mathrm{RMSE}_{90}(\mathbf{e}_s)
+ 0.10\,\mathrm{median}(\mathbf{e}_s) \\
&+ 0.05\,P_{90}(\mathbf{e}_s)
+ J_{\mathrm{short}}(s)
+ J_{\mathrm{prior}}(\boldsymbol{\theta}).
\end{aligned}
\label{eq:segment_calibration_objective}
\end{equation}

Here, $\mathrm{RMSE}_{90}$ denotes the RMSE after removing the largest 10\% of residuals. 
The term $J_{\mathrm{short}}$ penalizes candidate solutions that produce too few valid radar--LiDAR associations, while $J_{\mathrm{prior}}$ is a weak quadratic regularization term used to discourage excessive spatial or temporal corrections.

The final calibration used the five calibration runs to produce a joint multi-run objective. Objectives were balanced per run to avoid runs with high sample counts dominating the solution. For each run, approach and depart segments were first averaged and the resulting run objectives were then averaged over the set:

\begin{equation}
J_{\mathrm{joint}}(\boldsymbol{\theta})
=
\frac{1}{M}
\sum_{m=1}^{M}
\frac{1}{|\mathcal{S}_m|}
\sum_{s\in\mathcal{S}_m}
J_s(\boldsymbol{\theta}),
\qquad M=5.
\label{eq:calibration_objective}
\end{equation}

To find the optimized solution, we used a global population-based differential evolution search followed by a local Powell refinement. The selected solution was the lower-objective result between the differential evolution and the Powell candidate:

\begin{equation}
\boldsymbol{\theta}^{\star}
=
\arg\min_{\boldsymbol{\theta}\in\Omega}
J_{\mathrm{joint}}(\boldsymbol{\theta}),
\label{eq:calibration_optimization}
\end{equation}

where $\Omega$ denotes the search space for the calibration parameters.

The target vehicle GNSS trajectory was timestamp-aligned to the CV time base using camera-frame matching and transformed into the radar-centric Cartesian frame for evaluation.

After these calibration and preprocessing steps, radar detections, LiDAR detections, fused tracks, and the target vehicle's reference trajectory are all represented in the same two-dimensional metric coordinate frame and time base. This ensures that subsequent association, filtering, and evaluation are performed in a common coordinate frame and time base.

\subsection{Evaluation Procedure} \label{evaluation}

The parameters used in the final evaluation were selected through grid-based fine-tuning on a separately reserved calibration dataset. The calibration data was not included in the evaluation results. The final selected values are listed in Table~\ref{tab:final_tuned_parameters}.

\begin{table}[h]
\centering
\caption{Final parameter values selected after tuning. Parameters not used by a method are marked with --.}
\label{tab:final_tuned_parameters}
\footnotesize
\begin{tabular}{lcccc}
\hline
\textbf{Parameter} & \textbf{LiDAR} & \textbf{Radar} & \textbf{SEKF} & \textbf{AEKF} \\
\hline
Association gate $\gamma_{\mathrm{assoc}}$ 
    & 24 & 44 & 32 & 36 \\

Process noise $\sigma_a$ [m/s$^2$] 
    & 3.5 & 8.0 & 4.5 & 5.0 \\

Yaw-rate noise $\sigma_{\dot{\psi}}$ [rad/s] 
    & 0.65 & 0.85 & 0.95 & 0.95 \\

Radar position noise $\sigma_r$ [m] 
    & -- & 1.0 & 9.0 & 9.0 \\

LiDAR position noise $\sigma_l$ [m] 
    & 0.5 & -- & 0.4 & 0.5 \\

\hline
\end{tabular}
\end{table}

Evaluation was performed as a separate post-processing step after fusion. Thus, the fusion methods did not use the target vehicle identity, and the produced tracks were matched to the target vehicle reference trajectory only during evaluation.

\subsubsection{Track-to-Reference Matching}

For each reference sample, an estimated track is selected based on Euclidean distance to the target vehicle's reference position. A match is accepted only if it passes the evaluation gate $g_\mathrm{eval}=5.0$~m, which was selected to avoid artificially lowering the localization RMSE of the radar-only method.

To reduce artificial track switching during evaluation, sticky matching is used. After an initial track match, the evaluator remains linked to the same track ID as long as the track remains within the evaluation gate. If the linked track is unavailable for multiple consecutive samples, reacquisition is allowed after a fixed number of missed samples. In the final configuration, the initial gate and reacquisition gate are both $5.0$~m, and reacquisition is allowed after 10 missed samples which equates to one second at the radar sampling frequency.

\subsubsection{LiDAR Refresh Rate Ablation}

To simulate limited CV broadcasting or sensing capability, LiDAR measurements were made available to the fusion filter at reduced update rates. This ablation evaluates how sparse CV observations affect fusion performance and when radar updates improve over LiDAR-only localization.

This ablation was performed by limiting how frequently LiDAR measurements were made available to the fusion filter. For this ablation, the methods were tuned with the reduced LiDAR update rate of $1~\mathrm{Hz}$. The final tuned parameters for this ablation are shown in Table~\ref{tab:final_tuned_1hz_parameters}.

\begin{table}[h]
\centering
\caption{Final parameter values selected after tuning with LiDAR at $1~\mathrm{Hz}$. Parameters not used by a method are marked with --.}
\label{tab:final_tuned_1hz_parameters}
\footnotesize
\begin{tabular}{lcccc}
\hline
\textbf{Parameter} & \textbf{LiDAR} & \textbf{SEKF} & \textbf{AEKF} \\
\hline
Association gate $\gamma_{\mathrm{assoc}}$ 
    & 24 & 24 & 46 \\

Process noise $\sigma_a$ [m/s$^2$] 
    & 8.0 & 10.5 & 7.5 \\

Yaw-rate noise $\sigma_{\dot{\psi}}$ [rad/s] 
    & 0.65 & 0.6 & 0.9 \\

Radar position noise $\sigma_r$ [m] 
    & -- & 10.0 & 8.0 \\

LiDAR position noise $\sigma_l$ [m] 
    & 0.8 & 0.6 & 0.5 \\

\hline
\end{tabular}
\end{table}

\subsubsection{Simulated LiDAR Occlusions}

To evaluate robustness to loss of LiDAR observations of the target vehicle, Lidar occlusions were generated before fusion. For each case, LiDAR detections within a $2.5~\mathrm{m}$ gate of the target vehicle reference position were removed during a fixed time window. The tested occlusion durations were $3$, $5$, $7$, and $10~\mathrm{s}$. For each duration, the occlusion window was placed at three relative start positions, corresponding to $25\%$, $50\%$, and $75\%$ of the segment duration. Fusion and evaluation were then performed using the altered LiDAR data. This ablation tests whether radar updates can compensate when target-specific LiDAR detections are temporarily unavailable.

\subsubsection{Motion-Conditioned Evaluation}

In addition to localization metrics, performance is evaluated separately for different target-vehicle motion states. The target vehicle reference speed is computed from the horizontal GNSS velocity components. Samples are divided into three motion bins: stationary, slow, and moving.  RMSE, MAE (mean average error), and match rate are then computed separately for each bin. The number of samples in each bin is shown in Table~\ref{tab:motion_bin_sample_counts}.

\begin{table}[ht]
\centering
\caption{Distribution of reference samples across motion bins.}
\label{tab:motion_bin_sample_counts}
\footnotesize
\begin{tabular}{lcc}
\hline
\textbf{Motion bin} & \textbf{Speed range} & \textbf{Samples} \\
\hline
Stationary & $v_\mathrm{ref}<0.3$~m/s & 27845 \\
Slow & $0.3 \leq v_\mathrm{ref}<2.0$~m/s & 5163 \\
Moving & $v_\mathrm{ref}\geq2.0$~m/s & 27815 \\
\hline
Total &  & 60823 \\
\hline
\end{tabular}
\end{table}

\section{Results}

The evaluated methods were compared against the target vehicle GNSS reference trajectory. Localization accuracy is reported using RMSE and MAE over averaged segment results, while match rate and drop rate quantify the availability of valid estimates. The comparison includes LiDAR-only EKF, radar-only EKF, SEKF, and AEKF.

\subsection{Overall Localization Performance}

Table~\ref{tab:main_results} summarizes the overall localization performance of the methods across validation segments. LiDAR-only localization provides a strong baseline, while radar-only localization has considerably lower match rate and higher error. The fusion methods remain close to the LiDAR-only baseline, with AEKF giving the lowest overall RMSE and MAE by a small margin.

\begin{table}[h]
\centering
\caption{Overall performance comparison of evaluated methods across all validation segments.}
\label{tab:main_results}
\footnotesize
\begin{tabular}{lcccc}
\hline
\textbf{Method} & \textbf{RMSE [m] (std)} & \textbf{MAE [m]} & \textbf{Match rate} & \textbf{Drop rate} \\
\hline
LiDAR & 0.84 (0.30) & 0.74 & 0.983 & 0.017 \\
Radar & 3.27 (0.65) & 3.12 & 0.215 & 0.785 \\
SEKF & 0.84 (0.29) & 0.74 & 0.982 & 0.018 \\
AEKF & \textbf{0.81 (0.29)} & \textbf{0.73} & \textbf{0.984} & \textbf{0.016} \\
\hline
\end{tabular}
\end{table}

The minuscule differences between LiDAR-only, SEKF, and AEKF indicate that the LiDAR detections dominate the full-rate localization. The radar-only result confirms that the roadside radar alone is not enough for accurate continuous localization in this experiment.

Table~\ref{tab:segment_results} separates the results by approach and depart segments. The by segment comparison is used to showcase the variation between the segments.

\begin{table}[h]
\centering
\caption{Performance comparison of evaluated methods across all validation segments.}
\label{tab:segment_results}
\scriptsize
\begin{tabular}{llccc}
\hline
\textbf{Method} & \textbf{Segment} & \textbf{RMSE [m] (std)} & \textbf{MAE [m]} & \textbf{Match rate} \\
\hline
\multirow{2}{*}{LiDAR} 
 & Approach & \textbf{0.90 (0.36)} & \textbf{0.82} & \textbf{0.969} \\
 & Depart   & 0.78 (0.23) & 0.67 & 0.995 \\
\hline
\multirow{2}{*}{Radar} 
 & Approach & 2.93 (0.57) & 2.76 & 0.379 \\
 & Depart   & 3.61 (0.56) & 3.48 & 0.067 \\
\hline
\multirow{2}{*}{SEKF} 
 & Approach & 0.92 (0.37) & 0.82 & 0.969 \\
 & Depart   & 0.78 (0.19) & 0.66 & 0.994 \\
\hline
\multirow{2}{*}{AEKF} 
 & Approach & \textbf{0.90 (0.36)} & \textbf{0.82} & \textbf{0.969} \\
 & Depart   & \textbf{0.73 (0.20)} & \textbf{0.65} & \textbf{0.997} \\
\hline
\end{tabular}
\end{table}

The segmented results show similar relative behavior between the methods, although the absolute error varies between segment types. This indicates that the comparison is affected by segment geometry and visibility, but that the relative performance of the methods remains generally consistent.

\begin{figure*}[h]
    \centering
    \includegraphics[width=\textwidth]{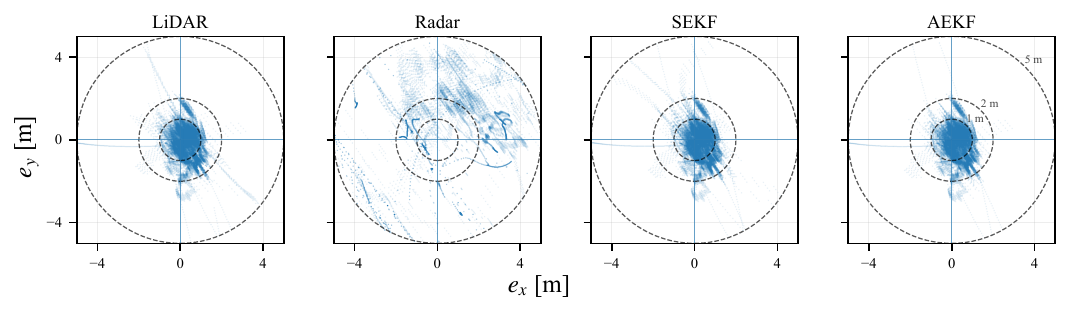}
    \caption{Pooled localization error clouds relative to the target vehicle's reference trajectory for the LiDAR-only, radar-only, SEKF, and AEKF methods. The dashed circles indicate radial error magnitudes of $1~\mathrm{m}$, $2~\mathrm{m}$, and $5~\mathrm{m}$.}
    \label{fig:error_cloud_methods_grid}
\end{figure*}

Figure~\ref{fig:error_cloud_methods_grid} shows the pooled localization error distributions for the evaluated methods. The radar-only EKF shows a substantially wider and more scattered error distribution. In contrast, the LiDAR-only EKF, SEKF, and AEKF methods remain mainly around the origin. This further shows that at the full LiDAR update rate the localization performance is dominated by the LiDAR observations.

\subsection{Robustness and Sensitivity Analysis}

\subsubsection{LiDAR Sparsity Analysis}

Table~\ref{tab:lidar_refresh_rate_ablation} evaluates the effect of reducing the LiDAR update rate. The error remains relatively stable at $5~\mathrm{Hz}$ and $3~\mathrm{Hz}$, but increases more clearly at $2~\mathrm{Hz}$ and below. The comparison between the full-rate tune and the $1~\mathrm{Hz}$ tune shows that retuning can improve sparse-rate performance, although the benefit is not uniform across methods or refresh rates.

\begin{table}[h]
\centering
\caption{Performance of the compared methods under varying LiDAR refresh rates. Values in parentheses next to RMSE denote the relative RMSE change compared with the best-performing method at the same refresh rate.}
\label{tab:lidar_refresh_rate_ablation}
\footnotesize
\begin{tabular}{@{}lll r@{\,}l cc@{}}
\hline
\textbf{Rate} & \textbf{Method} & \textbf{Tune} &
\multicolumn{2}{c}{\textbf{RMSE [m] ($\Delta$)}} &
\textbf{MAE [m]} & \textbf{Match} \\
\hline
\multirow{6}{*}{$5~\mathrm{Hz}$} & \multirow{2}{*}{LiDAR} & Full & 0.84 & (+1.5\%) & 0.74 & \textbf{0.981} \\
 &  & $1~\mathrm{Hz}$ & 0.83 & (+0.9\%) & 0.74 & 0.981 \\
 & \multirow{2}{*}{SEKF} & Full & 0.86 & (+3.8\%) & 0.75 & 0.979 \\
 &  & $1~\mathrm{Hz}$ & 0.85 & (+3.0\%) & 0.75 & 0.980 \\
 & \multirow{2}{*}{AEKF} & Full & \textbf{0.82} & \textbf{(+0.0\%)} & \textbf{0.74} & 0.981 \\
 &  & $1~\mathrm{Hz}$ & 0.84 & (+2.4\%) & 0.75 & 0.980 \\
\hline
\multirow{6}{*}{$3~\mathrm{Hz}$} & \multirow{2}{*}{LiDAR} & Full & 0.90 & (+1.7\%) & 0.78 & 0.977 \\
 &  & $1~\mathrm{Hz}$ & \textbf{0.89} & \textbf{(+0.0\%)} & \textbf{0.77} & 0.977 \\
 & \multirow{2}{*}{SEKF} & Full & 0.90 & (+1.9\%) & 0.78 & \textbf{0.977} \\
 &  & $1~\mathrm{Hz}$ & 0.90 & (+1.5\%) & 0.78 & 0.977 \\
 & \multirow{2}{*}{AEKF} & Full & 0.89 & (+0.9\%) & 0.78 & 0.977 \\
 &  & $1~\mathrm{Hz}$ & 0.91 & (+2.9\%) & 0.78 & 0.976 \\
\hline
\multirow{6}{*}{$2~\mathrm{Hz}$} & \multirow{2}{*}{LiDAR} & Full & 1.20 & (+18.7\%) & 0.95 & 0.946 \\
 &  & $1~\mathrm{Hz}$ & \textbf{1.01} & \textbf{(+0.0\%)} & \textbf{0.84} & \textbf{0.964} \\
 & \multirow{2}{*}{SEKF} & Full & 1.12 & (+11.0\%) & 0.93 & 0.952 \\
 &  & $1~\mathrm{Hz}$ & 1.04 & (+3.4\%) & 0.86 & 0.962 \\
 & \multirow{2}{*}{AEKF} & Full & 1.03 & (+1.9\%) & 0.85 & 0.957 \\
 &  & $1~\mathrm{Hz}$ & 1.07 & (+5.7\%) & 0.88 & 0.952 \\
\hline
\multirow{6}{*}{$1~\mathrm{Hz}$} & \multirow{2}{*}{LiDAR} & Full & 1.58 & (+4.9\%) & 1.23 & \textbf{0.901} \\
 &  & $1~\mathrm{Hz}$ & 1.59 & (+5.5\%) & 1.26 & 0.875 \\
 & \multirow{2}{*}{SEKF} & Full & 1.68 & (+11.3\%) & 1.33 & 0.881 \\
 &  & $1~\mathrm{Hz}$ & \textbf{1.51} & \textbf{(+0.0\%)} & \textbf{1.18} & 0.888 \\
 & \multirow{2}{*}{AEKF} & Full & 1.55 & (+2.8\%) & 1.21 & 0.887 \\
 &  & $1~\mathrm{Hz}$ & 1.53 & (+1.1\%) & 1.19 & 0.878 \\
\hline
\multirow{6}{*}{$0.5~\mathrm{Hz}$} & \multirow{2}{*}{LiDAR} & Full & 2.22 & (+6.1\%) & 1.78 & 0.728 \\
 &  & $1~\mathrm{Hz}$ & \textbf{2.09} & \textbf{(+0.0\%)} & \textbf{1.65} & 0.741 \\
 & \multirow{2}{*}{SEKF} & Full & 2.15 & (+3.0\%) & 1.73 & \textbf{0.767} \\
 &  & $1~\mathrm{Hz}$ & 2.40 & (+14.8\%) & 1.98 & 0.734 \\
 & \multirow{2}{*}{AEKF} & Full & 2.19 & (+4.6\%) & 1.77 & 0.765 \\
 &  & $1~\mathrm{Hz}$ & 2.21 & (+5.9\%) & 1.77 & 0.747 \\
\hline

\end{tabular}
\end{table}

At lower refresh rates, the effect of tuning becomes more noticeable, particularly around $1-2~\mathrm{Hz}$. At $0.5~\mathrm{Hz}$, all methods show substantial degradation, indicating that the prediction interval becomes too long for reliable localization without more frequent LiDAR corrections.

\subsubsection{Simulated LiDAR Occlusions}

Table~\ref{tab:lidar_occlusion_ablation} evaluates robustness when LiDAR observations are artificially reduced. This isolates the contribution of radar updates during periods where the CV is unable to detect the target vehicle.

\begin{table}[h]
\centering
\caption{Effect of simulated LiDAR occlusions. Metrics are computed only during the occlusion interval and averaged over occlusion start positions. Values in parentheses denote the change relative to the corresponding non-occluded same-window reference. Match-rate changes are reported as absolute proportions.}
\label{tab:lidar_occlusion_ablation}
\footnotesize
\begin{tabular}{llccc}
\hline
\textbf{Occlusion} & \textbf{Method} & \textbf{RMSE [m]} & \textbf{MAE [m]} & \textbf{Match rate} \\
\hline
\multirow{3}{*}{$3~\mathrm{s}$} & LiDAR & \textbf{1.34 (+0.64)} & \textbf{1.17 (+0.48)} & \textbf{0.904 (-0.087)} \\
 & SEKF & 1.46 (+0.74) & 1.26 (+0.56) & 0.899 (-0.091) \\
 & AEKF & 1.43 (+0.72) & 1.23 (+0.53) & 0.852 (-0.140) \\
\hline
\multirow{3}{*}{$5~\mathrm{s}$} & LiDAR & \textbf{1.91 (+1.21)} & \textbf{1.59 (+0.91)} & \textbf{0.774 (-0.219)} \\
 & SEKF & 2.03 (+1.30) & 1.70 (+1.00) & 0.767 (-0.225) \\
 & AEKF & 2.07 (+1.36) & 1.71 (+1.02) & 0.710 (-0.283) \\
\hline
\multirow{3}{*}{$7~\mathrm{s}$} & LiDAR & \textbf{2.42 (+1.70)} & \textbf{1.97 (+1.28)} & 0.692 (-0.302) \\
 & SEKF & 2.49 (+1.78) & 2.10 (+1.41) & \textbf{0.724 (-0.269)} \\
 & AEKF & 2.48 (+1.78) & 2.09 (+1.41) & 0.682 (-0.312) \\
\hline
\multirow{3}{*}{$10~\mathrm{s}$} & LiDAR & \textbf{2.81 (+2.10)} & \textbf{2.38 (+1.70)} & 0.647 (-0.347) \\
 & SEKF & 2.92 (+2.21) & 2.53 (+1.85) & \textbf{0.656 (-0.338)} \\
 & AEKF & 2.95 (+2.26) & 2.53 (+1.86) & 0.631 (-0.364) \\
\hline

\end{tabular}
\end{table}

The LiDAR-only EKF achieves the lowest RMSE and MAE for all tested occlusion durations, indicating that radar updates do not improve localization accuracy when targeted occlusion areas surround the target vehicle. However, SEKF maintains a slightly higher match rate for the longer $7~s$ and $10~s$ occlusions, suggesting that radar observations can help preserve track availability even when they do not reduce localization error.

\subsubsection{Motion-conditioned Analysis}

Table~\ref{tab:motion_bin_rmse_fulltune} reports localization RMSE separately for the motion regimes defined in Table~\ref{tab:motion_bin_sample_counts}. The comparison shows how reduced LiDAR update rates affect localization accuracy under different target-vehicle motion conditions.

\begin{table}[h]
\centering
\caption{Motion-conditioned localization RMSE in meters under varying LiDAR refresh rates using the full-rate tuned configurations.}
\label{tab:motion_bin_rmse_fulltune}
\footnotesize
\begin{tabular}{llccc}
\hline
\textbf{Rate} & \textbf{Motion bin} & \textbf{LiDAR} & \textbf{SEKF} & \textbf{AEKF} \\
\hline
\multirow{3}{*}{10 Hz} & Stationary & \textbf{0.60} & \textbf{0.60} & \textbf{0.60} \\
 & Slow & \textbf{0.72} & \textbf{0.72} & \textbf{0.72} \\
 & Moving & 0.92 & 0.94 & \textbf{0.88} \\
\hline
\multirow{3}{*}{5 Hz} & Stationary & 0.61 & 0.61 & \textbf{0.60} \\
 & Slow & 0.73 & \textbf{0.72} & 0.73 \\
 & Moving & 0.92 & 0.97 & \textbf{0.90} \\
\hline
\multirow{3}{*}{3 Hz} & Stationary & \textbf{0.61} & 0.62 & \textbf{0.61} \\
 & Slow & 0.74 & 0.74 & \textbf{0.72} \\
 & Moving & 1.02 & 1.01 & \textbf{1.00} \\
\hline
\multirow{3}{*}{2 Hz} & Stationary & \textbf{0.62} & 0.67 & \textbf{0.62} \\
 & Slow & 0.77 & 0.90 & \textbf{0.76} \\
 & Moving & 1.51 & 1.32 & \textbf{1.21} \\
\hline
\multirow{3}{*}{1 Hz} & Stationary & 0.74 & 0.85 & \textbf{0.63} \\
 & Slow & \textbf{1.15} & 1.30 & 1.27 \\
 & Moving & 2.14 & 2.22 & \textbf{2.11} \\
\hline
\multirow{3}{*}{0.5 Hz} & Stationary & 1.43 & \textbf{1.10} & 1.36 \\
 & Slow & 2.23 & \textbf{1.94} & 2.01 \\
 & Moving & 2.89 & \textbf{2.80} & 2.86 \\
\hline
\end{tabular}
\end{table}

As expected, the stationary regime is less affected by reduced LiDAR update rate than the moving regime, although degradation is still visible at the lowest refresh rate. The moving regime shows the clearest increase in error as the update interval grows, indicating that sparse LiDAR observations are most problematic when the target position changes substantially between updates. AEKF gives the lowest RMSE in most motion bins from $10~\mathrm{Hz}$ to $1~\mathrm{Hz}$, while SEKF performs best at $0.5~\mathrm{Hz}$.

\section{Discussion}

\subsection{Overall Interpretation of the Fusion Results}

Assessment of the decision-level fusion indicates that some performance can be gained when highly accurate LiDAR data is fused with lower quality roadside radar data. As Table~\ref{tab:main_results} shows, AEKF performed mildly better than the LiDAR-only EKF in terms of localization accuracy. SEKF and LiDAR-only EKF performed nearly identically when full rate data was available. In comparison, radar-only EKF had around four times higher localization error and had a very low match rate.

The results also indicate that the fusion methods mostly inherit the performance from the LiDAR, but that does not render fusion useless. The limited improvement from fusion mainly shows that with a strong full-rate LiDAR baseline there is limited room for improvement, especially with less accurate and inconsistent radar observations.

\subsection{Effect of Sensor Availability}

Reducing the LiDAR update rate degrades the localization performance across all methods, however, the degradation stays modest down to $3~\mathrm{Hz}$, where only approximately a 10 \% RMSE increase is noted as shown in Table~\ref{tab:lidar_refresh_rate_ablation}. The degradation ramps up under $2~\mathrm{Hz}$ and then the update rate specific tuning shows some performance gain. Overall the fusion performance is still closely matched with the LiDAR-only EKF, showing that the radar does not consistently improve performance. However, as with the full-rate comparison, some small performance can be gained in specific scenarios.

By looking at the results produced with the synthetic occlusions of the target vehicle in Table~\ref{tab:lidar_occlusion_ablation}, we can see higher degradation when compared to the lowered update rates. This suggests that even a fairly short three second temporary occlusion can have a higher negative impact on localization accuracy than lowering the update rate to $2~\mathrm{Hz}$.

The results also show that the LiDAR-only EKF was able to produce highest localization performance at each occlusion duration. This further emphasizes the point that the radar data did not consistently improve the performance via fusion, however small track-availability benefit could be observed with SEKF at higher occlusion durations.

Overall, these results suggest that sharing object-level dynamic observations from CVs can be useful for infrastructure-side perception, even when the update rate is lower than nominal sensing capability. 

\subsection{Scenario-Dependent Performance}

Analyzing the data by segment shows that the scenarios impact the sensors differently. As the Table~\ref{tab:segment_results} show, the depart segment was more favorable for the LiDAR whereas radar-only EKF performed better in the approach segment. However, the main performance gain of AEKF over LiDAR-only was observed during the depart segment suggesting that even when the radar performs overall worse, it can still contribute to some performance gain.

The LiDAR-based performance stayed fairly symmetric between the segments, whereas the radar-only EKF showed more asymmetry indicating that the static mounting of the roadside radar makes it more susceptible to occlusions and measurement-geometry-related errors. The match rate being very low during the depart segment especially suggests that the conditions were very unfavorable for the radar.

The motion-conditioned results, as shown in Table~\ref{tab:motion_bin_rmse_fulltune}, suggest that the reduced LiDAR availability affects the fusion performance differently based on the target vehicle motion state. At full rate, highest performance gain in favor of AEKF was observed during the moving regime, showing that the fusion is most useful when the target is in motion. The same goes for all update rates, except for $0.5~\mathrm{Hz}$ where the SEKF outperformed AEKF. Stationary periods are less sensitive to lowered LiDAR update rates, and the LiDAR-only EKF was basically equal in terms of RMSE compared to the fusion methods.

\subsection{Implications for Decision-Level Fusion in ITS}

The results of this experiment suggest that the decision-level fusion framework is mainly beneficial from the perspective of infrastructure perception and, by extension, traffic management systems. From the CVs' perspective, the benefit over LiDAR-only EKF was proven limited because the LiDAR already provides accurate and reliable perception in good conditions. 

The results show that the benefit of fusion is scenario-dependent. Although radar fusion did not consistently improve localization accuracy, small gains were observed in specific cases, such as the depart segment and when the target vehicle was moving. This indicates that radar-LiDAR decision-level fusion is not a uniformly better alternative to LiDAR-only EKF localization, but rather a practical framework for combining heterogeneous observations when available sensors provide complementary data.

\subsection{Methodological Considerations}

The comparison between SEKF and AEKF shows that the exact fusion update formulation had a smaller effect than sensor availability and measurement quality. AEKF achieved the best overall localization accuracy, while SEKF provided slightly better track availability in the longer simulated LiDAR occlusions. This suggests that averaged updates can be beneficial when both sensors provide complementary data, where as sequential updates may be more robust when observations are intermittent.

Both fusion methods rely on predefined measurement covariance models for the observations. This makes the framework simple and lightweight, but it also means that the filter performance depends on the covariance values representing the true sensor uncertainty. In addition, the radar and LiDAR measurement errors are treated as conditionally independent given the true target state and the fixed calibration parameters. This is reasonable because the sensors are physically separate, use different sensing modalities, and have different dominant error sources. However, in practice, some errors may be correlated due to spatial calibration and transformation into the common radar-centric coordinate frame.

The evaluation depends on the selected track-reference association procedure. RMSE and MAE are computed only over matched estimates, while match rate describe the availability of the valid estimates. Therefore, localization accuracy and track availability should be interpreted together. A method with lower RMSE is not necessarily better in all cases if it also produces fewer valid matches, and a method with higher match rate may still have larger localization error.

Averaged segment-level metrics and pooled metrics emphasize different aspects of performance. Segment-level averages weight each segment equally, whereas pooled metrics are more influenced by longer segments and segments with more matched samples. Therefore, averaged segment results were used as the primary reporting format. This avoids giving disproportionate weight to segments extended by traffic-light waiting times.

The same evaluation gate and matching procedure were applied to all methods, supporting relative comparison. However, the absolute RMSE, MAE, and match-rate values depend on the selected gate size, and should therefore be interpreted under the stated evaluation protocol.

\subsection{Limitations of the Experiment and Future Work}

This study has several limitations that should be considered when interpreting the reported results. First, the framework was evaluated based on a limited real-world dataset collected in a specific road environment with a fixed roadside radar placement and a limited amount of CV and target vehicle arrangements. Thus, the absolute performance of the showcased methods should not be interpreted as limits for the general performance.

The dataset consists of $19$ segments and should be expanded in future work to include a wider range of traffic scenarios, sensor arrangements, infrastructure installations, and environmental conditions. This would also enable a more comprehensive assessment of the multi-object tracking capabilities of the framework and provide a more generalizable benchmark for future research.

Second, the sensing capabilities of the used sensors were imbalanced in favor of the CV LiDAR. It provided substantially more reliable vehicle detections and thus localization than the radar. This limited the potential contribution of the radar measurements in the fusion framework. Different radar placement, improved sensor coverage, or the use of multiple roadside sensors could change the relative contribution of infrastructure-based observations. Such extensions would also allow the framework to be evaluated in larger infrastructure-assisted perception systems.

Third, the reported localization metrics include uncertainty from spatial and temporal calibration, CV GNSS/INS localization, and target vehicle GNSS/INS localization. The target vehicle trajectory was kept independent of the tracking inputs, which supports unbiased evaluation, but it should still be interpreted as a reference trajectory rather than perfect ground truth. Future work could reduce these uncertainties through online spatial and temporal calibration, adaptive measurement covariance estimation, and more detailed uncertainty modeling.

Finally, this work lacks a direct numerical comparison against public benchmarks or previously published methods using the same sensing configuration. Although public V2X cooperative perception datasets are available, they do not align with the experimental setup or the evaluation criteria considered in this work. Instead, this paper evaluates the proposed methods against consistent internal baselines, including LiDAR-only EKF, radar-only EKF, SEKF, and AEKF. However, the release of the source code and dataset supports reproducibility and enables more direct benchmarking in subsequent work.

\section{Conclusion}



We introduced a multi-observer vehicle localization framework which combines detections from infrastructure and CV sources. The key novelty lies in the real-world evaluation of decision-level fusion between observations from a static roadside radar and a dynamic LiDAR-equipped CV for target-vehicle tracking in an urban traffic environment.

The framework was evaluated using data collected at an urban intersection in Helsinki, Finland, with a separately instrumented target vehicle used as the reference trajectory. We leveraged SEKF- and AEKF-based fusion strategies for tracking, and also added LiDAR-only EKF and radar-only EKF for the comparison. These methods were evaluated under nominal sensing conditions, reduced LiDAR update rates, simulated occlusions, and different target-vehicle motion states.  Further, the proposed architecture operates on compact vehicle-level detections rather than raw sensor data, making it suitable for practical ITS scenarios where bandwidth, computation, and asynchronous observation availability are important constraints.

Our results showed that under full LiDAR availability, less accurate and less consistent radar observations provided only limited additional improvement to the overall performance of the fusion. In contrast, the radar-only performance showed substantially lower localization accuracy and match rate. Small gains were nevertheless observed with AEKF when compared to LiDAR-only EKF. Overall, the fusion methods inherited most of their performance from the LiDAR.

Reduced LiDAR availability degraded the performance only modestly down to approximately $2$--$3~\mathrm{Hz}$, while stronger deterioration was observed below $2~\mathrm{Hz}$. This suggests that even object-level CV observations shared at reduced update rates can remain useful for infrastructure-side sensing in ITS applications.

The results therefore show that decision-level fusion is not automatically superior to a strong single-sensor baseline, but can provide scenario-dependent benefits when observation availability, sensor geometry, or imbalance in detection quality affect the localization performance.

Future work should evaluate the framework with larger datasets, multiple infrastructure sensors, and denser traffic and tracking scenarios. Further development should also consider online calibration, adaptive uncertainty modeling, and extension to key tracking targets such as vulnerable road users.

\bibliographystyle{IEEEtran}

{\footnotesize
\bibliography{References}
}
\begin{IEEEbiography} [{\includegraphics[width=1in,height=1.25in,clip,keepaspectratio]{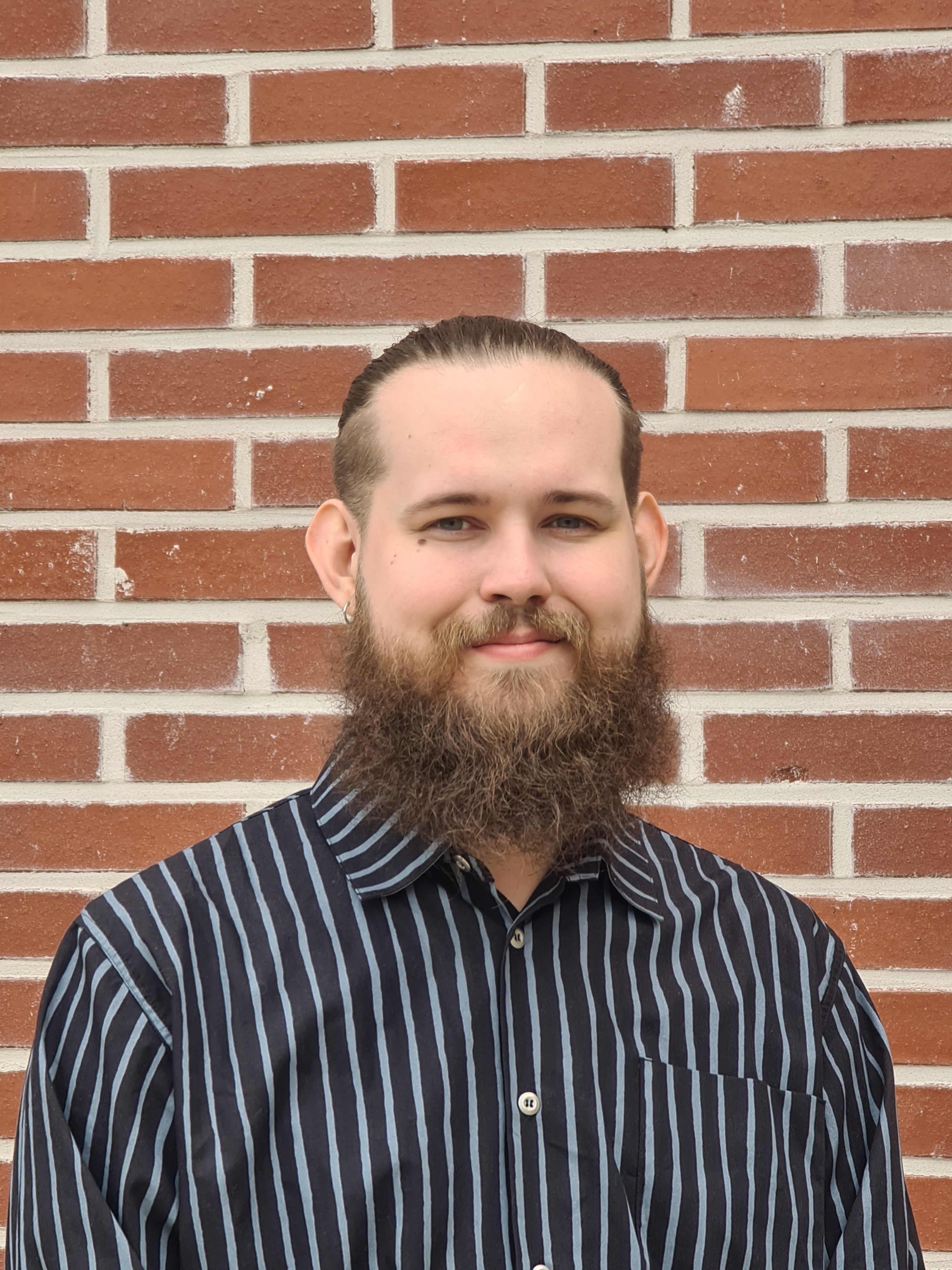}}]{Aleksi Pippuri} 
is currently a doctoral student at Aalto University, Finland. His doctoral research focuses on multi-observer fusion and collaborative sensing in mobile robotics. He received the B.Sc. and M.Sc degrees in mechanical engineering from Aalto University, Finland, in 2022 and 2024 respectively. His research interests include collaborative perception, intelligent transportation systems and autonomous driving. 
\end{IEEEbiography}

\begin{IEEEbiography} [{\includegraphics[width=1in,height=1.25in,clip,keepaspectratio]{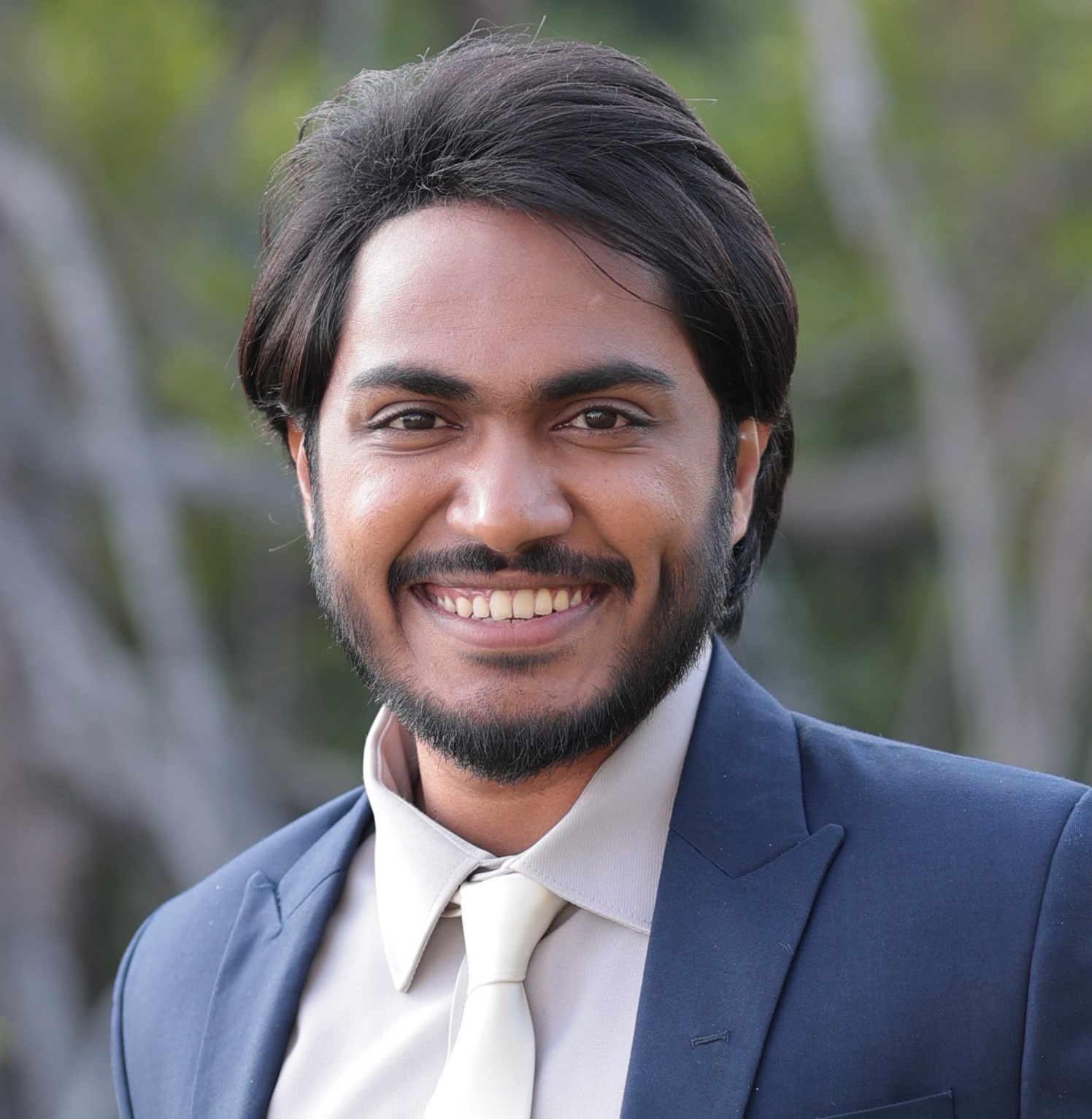}}]
{Nilusha Jayawickrama}
is currently a Postdoctoral Researcher with Aalto University, Finland, where he also received the D.Sc. degree in 2025. He has authored multiple peer-reviewed journal publications and his research interests include computer vision, machine learning, intelligent transportation systems, and mobile robotics. He has also been active in course development and teaching duties for autonomous vehicular control. He received his M.Sc. degree from Aalto University in 2020 and B.Sc. degree from Asian Institute of Technology in 2017.\end{IEEEbiography}

\begin{IEEEbiography} [{\includegraphics[width=1in,height=1.25in,clip,keepaspectratio]{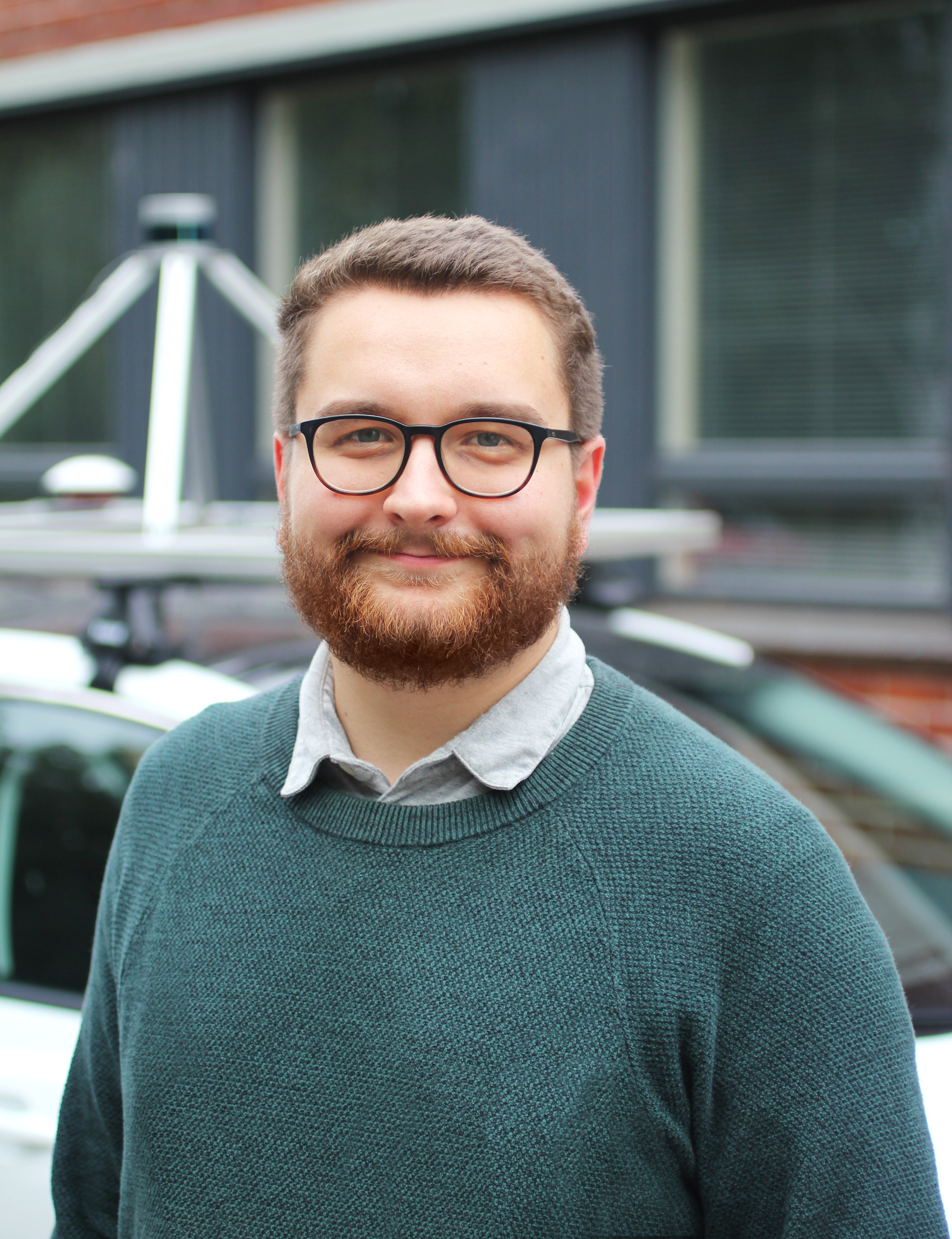}}]
{Risto Ojala}
received his BSc, MSc, and DSc degrees from Aalto University in 2019, 2021, and 2023, respectively. 
He served as a postdoctoral researcher at Aalto University from 2023 to 2024.
Currently, he continues at Aalto
as an assistant professor at the Autonomy \&
Mobility laboratory. He has authored several peer-reviewed journal publications, and his research interests focus on automated vehicles, mobile robotics,
computer vision, and machine learning. \end{IEEEbiography}

\end{document}